\documentclass[10pt,twocolumn,letterpaper]{article}

\usepackage{iccv}
\usepackage{times}
\usepackage{epsfig}
\usepackage{graphicx}
\usepackage{amsmath}
\usepackage{amssymb}

\DeclareMathOperator*{\argmin}{argmin}
\usepackage[noend]{algpseudocode}
\usepackage{algorithmicx,algorithm}
\usepackage{cite}
\usepackage{setspace}

\usepackage[breaklinks=true,bookmarks=false]{hyperref}

\iccvfinalcopy 

\def\iccvPaperID{1020} 

\ificcvfinal\fi

\makeatletter
\def\hlinew#1{%
  \noalign{\ifnum0=`}\fi\hrule \@height #1 \futurelet
   \reserved@a\@xhline}
\begin{document}

\title{Performance Guaranteed Network Acceleration via \\High-Order Residual Quantization}

\author{
Zefan Li$^1$, \hspace{0.4cm}Bingbing Ni$^1$, \hspace{0.4cm}Wenjun Zhang$^1$, \hspace{0.4cm}Xiaokang Yang$^1$,\hspace{0.4cm}Wen Gao$^2$\\
$^1$Shanghai Jiao Tong University, \hspace{0.4cm} $^2$Peking University\\
{\tt\small \{Leezf, nibingbing, zhangwenjun, xkyang\}@sjtu.edu.cn, \hspace{0.2cm}wgao@pku.edu.cn }
}

\maketitle
\thispagestyle{empty}

\begin{abstract}
Input binarization has shown to be an effective way for network acceleration. However, previous binarization scheme could be regarded as simple pixel-wise thresholding operations (i.e., order-one approximation) and suffers a big accuracy loss. In this paper, we propose a \emph{high-order} binarization scheme, which achieves more accurate approximation while still possesses the advantage of binary operation. In particular, the proposed scheme recursively performs residual quantization and yields a series of binary input images with decreasing magnitude scales. Accordingly, we propose high-order binary filtering and gradient propagation operations for both forward and backward computations. Theoretical analysis shows approximation error guarantee property of proposed method. Extensive experimental results demonstrate that the proposed scheme yields great recognition accuracy while being accelerated.

\end{abstract}

\section{Introduction}

Methods to accelerate learning and evaluation of deep network could be roughly divided into three groups. The simplest method is to perform network pruning (i.e., by rounding off near-zero connections) and re-train the pruned network structure~\cite{hassibi1993second,LeCun89,pratt1989comparing}. To achieve more structural compression rate, structural sparsity approximation techniques are later developed to morph larger sub-Networks into shallow ones~\cite{bucila2006model,DBLP:conf/bmvc/JaderbergVZ14,zhang2015efficient}. However, this type of method is not a general plug-in solution. Namely, for different networks with different network structures, expert knowledge is required to design the corresponding proper approximation network. Recently, a new set of solutions called \emph{network binarization} was proposed~\cite{courbariaux2015binaryconnect,courbariaux2016binarized,rastegari2016xnor}. The idea behind network binarization is simple: transform the floating weights of network as well as the corresponding forward or backward data flow to binary, therefore both computation and network storage could be reduced. For example, BinaryConnect-Network~\cite{courbariaux2015binaryconnect} shows great performance on datasets like CIFAR-10 and SVHN, but does not perform well enough on large-scale datasets (e.g., ImageNet). Binary-Weights-Network (BWN)~\cite{rastegari2016xnor} reduces the network storage by $\sim32\times$ and reach a state-of-art result on ImageNet dataset.

\begin{figure}[t]
\begin{center}
   \includegraphics[width=1\linewidth]{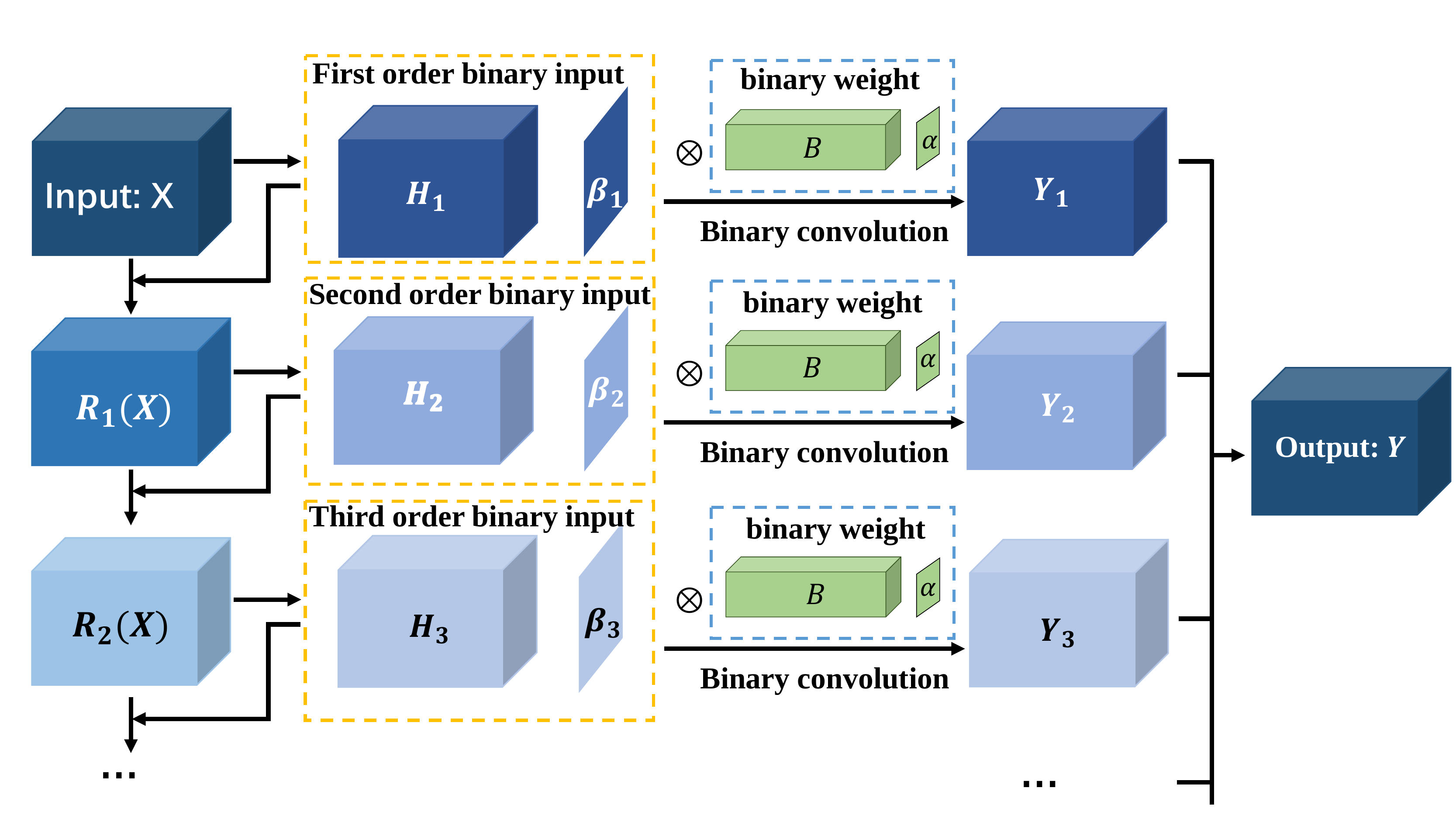}
\end{center}
   \caption{This figure shows how the High-Order Residual Quantization method operates on a common convolutional layer. $X$ is the input tensor. $R_i(X)$ is the $i$-th order residual (defined in Section~\ref{setction3.3}) of X. We use the first-order binary quantization $B$ and $\alpha$ of weight filter $W$. The final output $Y$ is the sum of outputs in different orders. In a Order-Two Residual Quantization, $Y=Y_1+Y_2$.}
\label{fig:long}
\label{fig:onecol}
\end{figure}

To further speed up network computation, input image is also binarized via thresholding operation. However, while network evaluation speed is reduced dramatically by a factor of $\sim58\times$, the recognition accuracy on ImageNet drop from $56.6\%$ to $27.9\%$ (BNN~\cite{courbariaux2016binarized}) and $44.2\%$ (XNOR~\cite{rastegari2016xnor}), due to large approximation error. Motivated by this limitation, in this work, we propose a High-Order Residual Quantization (HORQ) framework. The basic idea of this proposed framework is straightforward: previous input binarization operation, which simply performs positive and negative thresholding, could be considered as a very coarse quantization of floating numbers. In contrast, we propose a much more precise binary quantization method via recursive thresholding operation. Namely, after one time of thresholding operation, we could calculate the residual error and then perform a new round of thresholding operation to further approximate the residual. Thus, we could obtain a series of binary maps corresponding to different quantization scales. Based on these binary input tensors (stacked binary maps of different magnitude scales), we have developed efficient binary filtering operations for forward and backward computation. Experiments well demonstrate that our new proposed input binary quantization scheme not only outperforms the original XNOR-Networks~\cite{rastegari2016xnor}, but also possesses great speedup ratio. At the same time, theoretically, we provide error analysis for our approximation scheme.

The rest of this paper is organized as follows. Some related works are demonstrated and compared in Section~\ref{setction2}. In Section~\ref{setction3}, we propose the High-Order Residual Quantization method and HORQ-Net. Section~\ref{section4} covers the experiment part and analysis on storage and computation.
\section{Related Work}
\label{setction2}

Standard implementation of DCNNs is inefficient in memory storage and consumes considerable computational resources. Many works tried to accelerate and simplify DCNN. We divide these works into three categories:

\textbf{Parameter Pruning}
It is believed many deep learning models are over-parameterized with significant redundancy~\cite{Preciting13}. To simplify DCNN, a widely used method is to remove parameters with little information. An early method called weight decay~\cite{pratt1989comparing} is firstly used in pruning a network. OBD (Optimal Brain Damage~\cite{LeCun89}) provides a method using second-derivative information to remove the unimportant weights from a neural network under the assumption that the Hessian matrix of the problem is diagonal. OBS (Optimal Brain Surgeon~\cite{hassibi1993second}) furthers the idea of OBD and achieves a better experimental results. However, their methods need to compute the second-derivatives, which increases the computational complexity significantly.

Collins \etal~\cite{DBLP:journals/corr/CollinsK14} proposed a method using sparsity-inducing regularizer during the training of CNNs. Then Han \etal~\cite{han2015deep} proposed an approach to apply parameter pruning to a memory-efficient structure. Related approaches can be found in~\cite{DBLP:conf/bmvc/SrinivasB15} and~\cite{zhou2016less}. They firstly find similar neurons during the training process and then combine or remove these neurons. These methods are based on a pre-trained neural network. Our High-Order Residual Quantization method does not rely on a pre-trained network. Therefor, our method has an advantage of easy training.

\textbf{Model Compression}
Another approach to simplify neural networks is called model compression~\cite{bucila2006model}. The original idea of model compression is to train a compact artificial neural network to mimic a full-version pre-trained complex model. The full-version model is used to label the large unlabeled data set and the compact network is trained on this ensemble labeled data set. Compared with other compression methods, this method simply trains a network with fewer hidden units, thus the performance is limited.

Some other methods also consider the similar approach but develop many other skills. Jaderberg \etal~\cite{DBLP:conf/bmvc/JaderbergVZ14} proposed a method for accelerating convolutional networks in linear case and later, Zhang \etal~\cite{zhang2015efficient} proposed a method for accelerating convolutional networks in nonlinear case. These two methods minimize the reconstruction error of the responses (linear and nonlinear respectively) under the assumption that the convolutional filters can be low-rank approximated along certain dimensions. Methods in~\cite{DBLP:conf/cvpr/RigamontiSLF13},~\cite{denton2014exploiting} and~\cite{jaderberg2014speeding} approximate a weight filter with a set of separable smaller filters. These methods rely on the low-rank assumption and also need a pre-trained network. They are network-dependent. In contrast, our method is general and is a plug-in solution.

\textbf{Network Quantization}
This part is most related to our method. It is obvious that operations in high precision are much more time-consuming than those in binary values (eg. $+1,-1$). Training a DCNN with binary weights can significantly accelerate the computation since if the weight filters are replaced with binary values, the convolutional operation can be simply replaced by additions and subtractions. EPB (Expectation BackPropagation~\cite{DBLP:conf/nips/SoudryHM14}) shows that network with binary weights and binary activations is capable of achieving high performance. BC (BinaryConnect~\cite{courbariaux2015binaryconnect}) extends the idea of EBP and later, Courbariaux \etal~\cite{courbariaux2016binarized} proposed BinaryNet (BN) which is a further extension of BC. BC constrains weights to $+1$ and $-1$ and BN further constrains activations to $+1$ and $-1$ thus the input (except the first layer) and the output of each layer are all binary values. They both achieve a state-of-art result in small-scale data sets (eg. MNIST and CIFAR-10). According to Rastegari \etal~\cite{rastegari2016xnor}, BC and BN are not very successful on large-scale data set. Therefore, Rastegari \etal~\cite{rastegari2016xnor} proposed Binary-Weights-Networks (BWN) and XNOR-Networks, which use a different binary method compared with BC and BN. The most innovative point of their method is to compute the scaling factor. BWN uses binary weights and shows better performance than BC on ImageNet. XNOR, using both binary weights and binary input, further improves the efficiency of the computation. The accuracy of this network drops largely due to the information loss during input quantization. This inspires us to propose the High-Order Residual Quantization method, which reduces the information loss during quantization. We compare our HORQ method with XNOR and BN. Our method outperforms these previous methods.

\section{HORQ Network}
\label{setction3}

In this chapter, we propose a new binary quantization method named High-order Residual Quantization (HORQ) which realizes the binarization of both input and weights in a neural network. The most innovative point of HORQ is that we recursively make use of the residual (defined in Section~\ref{setction3.3}). Then we can obtain a series of binary inputs in different magnitude scales. We perform convolution operation on input in different scales and combine the results. This method manages to reduce the information loss during binary quantization.

We start with some notations. We use $\langle\mathcal{I},\mathcal{W},*\rangle$ to represent a convolutional neural network where $\mathcal{I}$ represents the set of input tensors, $\mathcal{W}$ represents the set of weight filters and $*$ represents the convolution operation. We use $I_l\in\mathcal{I}$ to represent the input tensor of the $l^{th}$ layer and $W_l\in\mathcal{W}$ to represent the weight filters of the $l^{th}$ layer. We use $c,w,h$ to represent \emph{channel, width and height} so that $I_l\in\mathbb{R}^{c_{in}\times w_{in}\times h_{in}}$ and $W_l\in\mathbb{R}^{c_{out}\times c_{in}\times w\times h}$.
\subsection{XNOR-Network Revisited}

In this section, we will briefly revisit the method proposed by Rastegari \etal~\cite{rastegari2016xnor}. They proposed two kinds of binary-neural-Networks named BWN and XNOR. BWN uses binary weights to speed up the computation. XNOR is based on BWN and realizes the binarization of input data in a convolutional layer. Here is a brief explanation of the quantization method used in XNOR and BWN:

Consider one convolution layer of the neural network with $I\in\mathcal{I}$ being the input tensor and $W\in\mathcal{W}$ being a weight filter. The core operation of this layer can be represented as $I*W$. The idea of BWN is to constrain a convolutional neural network with binary weights. Rastegari used $\alpha B$ to approximate $W$ where $\alpha\in\mathbb{R}^+$ is a scaling factor and $B\in\{-1,+1\}^{c\times w\times h}$ is a binary filter:
\begin{equation}
  I*W\approx(I\oplus B)\alpha
\end{equation}
Here, $\oplus$ represents a binary convolution operation with no multiplication. To find suitable $\alpha$ and $B$, Rastegari~\cite{rastegari2016xnor} solved the following optimization problem:
\begin{equation}\label{equaltion2}
  \alpha^*,B^*=\mathop{\argmin}_{\alpha,B}J(B,\alpha)=\mathop{\argmin}_{\alpha,B}\|W-\alpha B\|^2\\
\end{equation}
It is easy to find the solution to Equation~\ref{equaltion2}:
\begin{equation}\label{approxmation1}
  \left\{
  \begin{aligned}
      &B^*=sign(W) \\
      &\alpha^*=\frac{1}{n}\|W\|_{l_1}
  \end{aligned}
  \right.
\end{equation}
Using the optimal estimation (Equation~\ref{approxmation1}), one can train the CNN according to Algorithm~\ref{algorithm1} proposed by~\cite{rastegari2016xnor}.
\begin{algorithm}
\caption{Training an L-layers CNN with binary weights:} 
\label{algorithm1}
\hspace*{0.02in} {\bf Input:} 
A minibatch of inputs and targets $(I,Y)$, cost function $C(Y,\hat{Y})$, current weight $\mathcal{W}^t$ and current learning rate $\eta^t$\\
\hspace*{0.02in} {\bf Output:} 
Updated weight $\mathcal{W}^{t+1}$ and updated learning rate $\eta^{t+1}$
\begin{algorithmic}[1]
\State Binarizing weight filters: 
\For{$l=1$ to $L$} 
　　\For{$k=1$ to $c_{out}$} 
　　\State $A_{lk}=\frac{1}{n}\|\mathcal{W}^t_{lk}\|_{lk}$
    \State $B_{lk}=sign(\mathcal{W}^t_{lk})$
    \State $\widetilde{\mathcal{W}}_{lk}=A_{lk}B_{lk}$
    \EndFor
\EndFor
\State $\hat{Y}=$BinaryForward$(I,B,A)$
\State $\frac{\partial{C}}{\partial{\widetilde{\mathcal{W}}}}=$BinaryBackward$(\frac{\partial{C}}{\partial{\hat{Y}}},\widetilde{\mathcal{W}})$
\State $\mathcal{W}^{t+1}=$UpdateParameters$(\mathcal{W}^t,\frac{\partial{C}}{\partial{\widetilde{\mathcal{W}}}},\eta_t)$
\State
$\eta^{t+1}=$UpdateLearningrate$(\eta^t,t)$
\end{algorithmic}
\end{algorithm}

The idea of XNOR is based on BWN. BWN only replaces real value weights with binary values while XNOR is designed to replace real value inputs with binary values in addition to binary weights. XNOR uses $\beta H$ to approximate the input tensor $X$: $X\approx \beta H$ and they solve the following optimization problem:
\begin{equation}
  \alpha^*,B^*,\beta^*,H^*=\mathop{\argmin}_{\alpha,B,\beta,H}\|X\odot W-\alpha\beta H\odot B\|^2
\end{equation}
As showed in~\cite{rastegari2016xnor}, an approximate solution to this problem is:
\begin{equation}\label{XNORsolution}
  \left\{
  \begin{aligned}
      &\beta^*H^*=\frac{1}{n}\|X\|_{l_1}sign(H)\\
      &\alpha^*B^*=\frac{1}{n}\|W\|_{l_1}sign(W) \\
  \end{aligned}
  \right.
\end{equation}
We can use an algorithm similar to Algorithm~\ref{algorithm1} to train XNOR. More details about the training process of BWN and XNOR can be found in~\cite{rastegari2016xnor}. The experiments in~\cite{rastegari2016xnor} show that XNOR further accelerates the speed but the accuracy drops largely compared with BWN. Thus our purpose is to propose a improved neural networks of which both weights and inputs are binary values and the performance remains a relatively high level both in speed and accuracy. Based on this idea, in the next section, we propose the High-Order Residual Quantization method (HORQ).
\subsection{High-Order Residual Quantization }
\label{setction3.3}
In this section, we will explain the HORQ method to quantize the input of a convolutional layer.
Using $H^*$ and $\beta^*$ in Equation~\ref{XNORsolution} is not precise enough. Our HORQ method calculates the residual error and then performs a new round of thresholding operation to further approximate the residual. This binary approximation of the residual can be considered as a higher-order binary input. We can recursively perform the above operations and finally we can obtain a series of binary maps corresponding to different quantization scales. Based on these binary input tensors, we develop efficient binary filtering operations for forward and backward computation.

The input of a convolution layer is a 4-dimension tensor. If we reshape the input tensor and the corresponding weight filters into matrices, the convolution operation can be considered as a matrix multiplication. The process of tensor reshape will be demonstrated in Section~\ref{setction3.4}. Each elemental operation within the matrix production can be considered as a vector inner product operation. Thus we firstly consider the input as a vector:

Suppose there is an input vector $X\in\mathbb{R}^n$ and we quantize the $X$ following the process of XNOR:
 \begin{equation}\label{1st order}
   X\approx\beta_1H_1
 \end{equation}
 where $\beta_1\in\mathbb{R}$ and $H_1\in\{+1,-1\}^n$. We can get the result by solving the following optimization problem:
\begin{equation}
  \begin{aligned}
  \beta_1^*,H_1^*
  & =\mathop{\argmin}_{\beta_1,H_1}J(\beta_1,H_1)\\
  & =\mathop{\argmin}_{\beta_1,H_1}\|X-\beta_1H_1 \|^2
  \end{aligned}
\end{equation}
The analytical solution to this problem is:
 \begin{equation}\label{1st order solution}
  \left\{
  \begin{aligned}
      &H_1^*=sign(X) \\
      &\beta_1^*=\frac{1}{n}\|X\|_{l_1}
  \end{aligned}
  \right.
\end{equation}
%
Equation~\ref{1st order} can be considered as an order-one binary quantization(i.e., simple thresholding). Thus we can define the first-order residual tensor $R_1(X)$ by computing the difference between the real input and first-order binary quantization:
\begin{equation}
  R_1(X)=X-\beta_1H_1
\end{equation}
Since $\beta_1$ and $H_1$ can both be determined by $X$ from Equation~\ref{1st order solution}, $R_1(X)$ can also be determined by $X$. We can use $R_1(X)$ to represent the information loss due to approximation using Equation~\ref{1st order}. Notice that $R_1(X)$ is a real value tensor and we can further quantize $R_1(X)$ as follow:
\begin{equation}
  R_1(X)\approx \beta_2H_2
\end{equation}
where $\beta_2\in\mathbb{R}$, $H_2\in\{+1,-1\}^n$, then we can get the Order-Two Residual Quantization of the input:
\begin{equation}\label{2th order}
  X=\beta_1H_1+R_1(X)\approx \beta_1H_1+\beta_2H_2
\end{equation}
where $\beta_1,\beta_2$ are real value scalars and $H_1,H_2$ are binary value tensors. $\beta_1H_1$ is called the first-order binary input while $\beta_2H_2$ is called the second-order binary input. Using the similar way that we solve the Equation~\ref{1st order}, we can solve approximation problem of Equation~\ref{2th order}:\\
\noindent Firstly, we solve the corresponding optimization problem:
\begin{equation}\label{argmin2}
  \beta_2^*,H_2^*=\mathop{\argmin}_{\beta_2,H_2}\|R_1(X)-\beta_2H_2 \|^2\\
\end{equation}
and the solution to Problem~\ref{argmin2} is:
\begin{equation}
  \left\{
  \begin{aligned}
      &H_2^*=sign(R_1(X)) \\
      &\beta_2^*=\frac{1}{n}\|R_1(X)\|_{l_1}
  \end{aligned}
  \right.
\end{equation}
We can show that our binary approximation method of using Equation~\ref{2th order} is much better than the original method by using Equation~\ref{1st order} both theoretically and experimentally.

We can compare the information loss between these two binary approximation methods. Remember we define $R_1(X)$ as the residual tensor of approximation using Equation~\ref{1st order}. Then it's natural to define the residual tensor of approximation by Equation~\ref{2th order}:
\begin{equation}
\begin{aligned}
  R_2(X) &=X-\beta_1H_1-\beta_2H_2\\
         &=R_1(X)-\beta_2H_2
\end{aligned}
\end{equation}
Notice that $H_2^*$ and $\beta_2^*$ minimize $\|R_1(X)-\beta_2H_2\|^2$, therefore:
\begin{equation}\label{compare}
\begin{aligned}
     &\|R_2(X)|_{\beta_2=\beta_2^*,H_2=H_2^*}\|^2\\
    =&\|(R_1(X)-\beta_2H_2)\|^2|_{\beta_2=\beta_2^*,H_2=H_2^*}\\
    =&\|(R_1(X)-\beta_2H_2)\|^2_{min}\\
    \leqslant &\|(R_1(X)-\beta_2H_2)\|^2|_{\beta_2=0}\\
    =&\|R_1(X)\|^2\\
\end{aligned}
\end{equation}
Thus if we use the $L2-norm$ of the residual tensor to represent the information loss, from the above derivation, we can prove that our Order-Two Residual Quantization by Equation~\ref{2th order} reduces the information loss compared with the approximation using Equation~\ref{1st order} in~\cite{rastegari2016xnor}.

It's straightforward to develop the Order-Two Residual Quantization using Equation~\ref{2th order} into a Order-K Residual Quantization:
\begin{equation}\label{kth order}
  X\approx \sum_{i=1}^K\beta_iH_i
\end{equation}
where
\begin{equation}
  \left\{
  \begin{aligned}
      &R_0(X)=X\\
      &R_{i-1}(X)=X-\sum_{j=1}^{i-1}\beta_jH_j &i=2,3,...,K\\
      &H_{i}=sign(R_{i-1}(X)) &i=1,2,...,K\\
      &\beta_{i}=\frac{1}{n}\|R_{i-1}(X)\|_{l_1} &i=1,2,...,K
  \end{aligned}
  \right.
\end{equation}
We can recursively calculate the residual tensor to get a higher-order input. In fact, if the order becomes higher, the information loss will be more less, while the computational cost will also increase. We find that Order-Two and Order -Three residual quantization are good enough to approximate the input in terms of information loss. In the next section, we will introduce the HORQ network using our Order-Two Residual quantization method.
\subsection{The HORQ Network}
\label{setction3.4}

In this section, we proposed HORQ-Net which takes the HORQ binary input and performs high order binary filtering for forward and backward computation. As for the convolutional layer, suppose the input $X\in\mathbb{R}^{c_{in}\times w_{in}\times h_{in}}$ and the convolutional filter $W\in\mathbb{R}^{c_{out}\times c_{in}\times w\times h}$ of this convolution layer are two tensors. Then if we reshape the input tensor and weight tensor into two matrices respectively, the convolution operation can be considered as a matrix multiplication.

\textbf{Tensor Reshape}
To reshape the weight tensor $W$, we can straighten each filter to a vector shape of $1\times(c_{in}\times w\times h)$. There are $c_{out}$ filters thus the weight tensor $W$ is reshaped to a matrix $W_r$ shape of $c_{out}\times(c_{in}\times w\times h)$. If we use $Y$ to denote the output of the convolution layer $\langle X,W,*\rangle$， then $Y\in\mathbb{R}^{c_{out}\times w_{out}\times h_{out}}$, where $w_{out}=(w_{in}+2*p-w)/s+1$ and  $h_{out}=(h_{in}+2*p-h)/s+1$, p and s represent the pad and stride parameter respectively. To reshape the input tensor $X$, we can straighten each sub-tensor in $X$ with the same size of a filter to a vector and combine these vectors to a matrix $X_r$. In fact, there are $w_{out}\times h_{out}$ sub-tensors in $X$, thus $X_r$ is in the shape of $(c_{in}\times w\times h)\times (w_{out}\times h_{out})$.
Then we can use a matrix production $Y_r=W_rX_r$ to replace the convolution operation between $X$ and $W$ where $Y_r$ is a matrix shape of $(c_{out})\times(w_{out}\times h_{out})$. Then we reshape $Y_r$ to $Y$ to complete the whole computation.
    \begin{figure}[h]
    \begin{center}
       \includegraphics[width=1\linewidth]{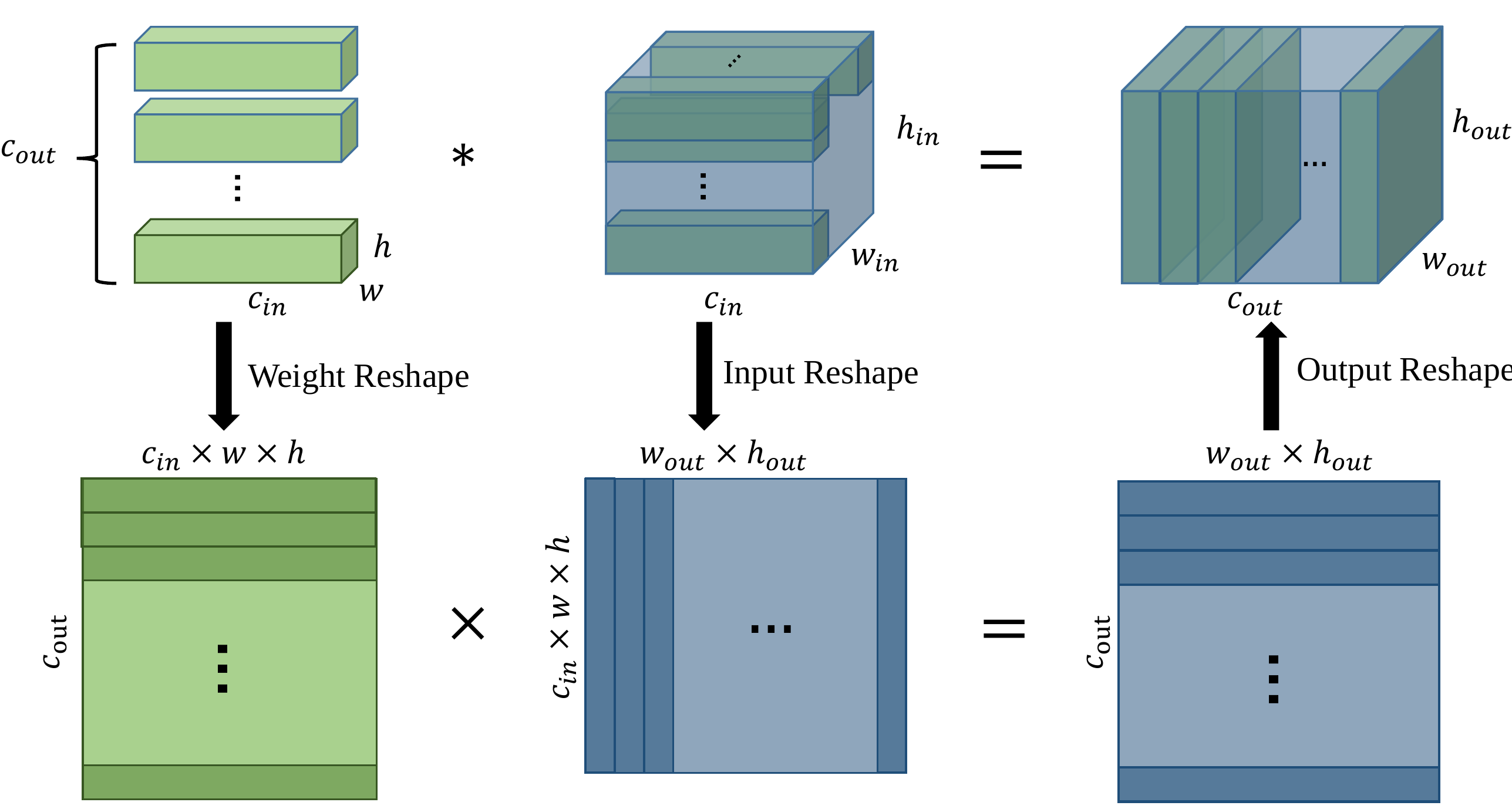}
    \end{center}
       \caption{This figure shows the tensor reshape process.}
    \label{fig:long}
    \label{fig:onecol}
    \end{figure}

        \begin{algorithm}
        \caption{OrderTwoBinaryConvolution$(X,W)$} 
        \label{SOBC}
        \hspace*{0.02in} {\bf Input:} 
        Input tensor $X\in\mathbb{R}^{c_{in}\times w_{in}\times h_{in}}$, Weight tensor $W\in\mathbb{R}^{c_{out}\times c_{in}\times w\times h}$ and convolutional parameters include pad and stride.\\
        \hspace*{0.02in} {\bf Output:} 
        The convolutional result $Y$ using  method of second-order binary approximation.
        \begin{algorithmic}[1]
        \State Reshape weight tensor and input tensor:
        \State $W_r$=ReshapeWeight($W$)
        \State $X_r$=ReshapeInput($X,W$)
        \State Binarizing weight matrix: 
            \For{$k=1$ to $c_{out}$}
            　　\State $A_{k}=\frac{1}{c_{n}\times w\times h}\|W_{r(k)}(t)\|_{l1}$
                \State $M_{k}=sign(W_{r(k)})$
                \State $\widetilde{W}_{r(k)}=A_{k}M_{k}$
            \EndFor
        \State Binarizing input matrix: 
            \For{$k=1$ to $w_{out}\times h_{out}$}
            　　\State $B_{1k}=\frac{1}{c_{n}\times w\times h}\|X_{r(k)}\|_{l1}$
                \State $N_{1k}=sign(X_{r(k)})$
                \State $R_1(X_{r(k)})=X_{r(k)}-B_{1k}N_{1k}$
                \State $B_{2k}=\frac{1}{c_{n}\times w\times h}\|R_1(X_{r(k)})\|_{l1}$
                \State $N_{2k}=sign(R_1(X_{r(k)}))$
                \State $\widetilde{X}_{r(k)}=B_{1k}N_{1k}+B_{2k}N_{2k}$
            \EndFor
        \State $Y_r=$BinaryProduction$(\widetilde{X}_{r(k)},\widetilde{W}_{r(k)})$
        \State $Y=$ReshapeOutput$(Y_r)$
        \end{algorithmic}
        \end{algorithm}

\textbf{Convolution Using Order-Two Residual Quantization}
After the Tensor Reshape process, we get the input $X_r$ and the weight $W_r$ in matrix form. In this part, we show how to use Order-Two Residual Quantization to compute the matrix production between $W_r$ and $X_r$. We firstly quantize the weight matrix $W_r$:
\begin{equation}
  W_{r(i)}\approx\alpha_i B_i\quad(i=1,2,...,c_{out})
\end{equation}
\begin{equation}
  \left\{
  \begin{aligned}
      &B_i=sign(W_{r(i)}) \\
      &\alpha=\frac{1}{c_{in}\times w\times h}\|W_{r(i)}\|_{l_1}
  \end{aligned}
  \right.
\end{equation}
where $W_{r(i)}$ is the $i$-th row of $W_r;\ W_{r(i)},B_i\in\mathbb{R}^{1\times (c_{in}\times w\times h)};\ \alpha\in\mathbb{R}$.

Then, we quantize the input matrix $X_r$ using Order-Two Residual Quantization:
\begin{equation}
  X_{r(i)}\approx\beta_{1(i)}H_{1(i)}+\beta_{2(i)}H_{2(i)}\quad(i=1,2,...,w_{out}\times h_{out})
\end{equation}
\begin{equation}
  \left\{
  \begin{aligned}
      &H_{1(i)}=sign(X_{r(i)}) \\
      &\beta_{1(i)}=\frac{1}{c_{in}\times w\times h}\|X_{r(i)}\|_{l_1}\\
      &R_1(X_{r(i)})=X_{r(i)}-\beta_{1(i)}H_{1(i)}\\
      &H_{2(i)}=sign(R_1(X_{r(i)})) \\
      &\beta_{2(i)}=\frac{1}{c_{in}\times w\times h}\|R_1(X_{r(i)})\|_{l_1}
  \end{aligned}
  \right.
\end{equation}
where $X_{r(i)}$ is the $i$-th column of $X_r;\ X_{r(i)},H_{1(i)},H_{2(i)}\in\mathbb{R}^{(c_{in}\times w\times h)\times 1};\ \beta_{1(i)},\beta_{2(i)}\in\mathbb{R}$. Thus we can compute the binary convolution via Algorithm~\ref{SOBC}.

\textbf{Training HORQ Network}
Algorithm~\ref{Myal} demonstrates the procedure for training a HORQ network using our Order-Two Residual Quantization method. The ordinary procedure includes Forward, Backward and Parameter-Update. We use the binary value of inputs and weights during the Forward and Backward process. For convenience, we only include convolution layers in the Forward process in Algorithm~\ref{Myal}. In fact, our High-Order Residual Quantization method can be easily applied to fully-connected layers because the fully connected layer only involves vector inner product and we can use our HORQ method directly without the Tensor Reshape process.

        \begin{algorithm}[h]
        \caption{Traning an L-layers HORQ network:} 
        \label{Myal}
        \hspace*{0.02in} {\bf Input:} 
        A minibatch of inputs and targets $(X,Y)$, cost function $L(Y,\hat{Y})$, current weight $\mathcal{W}(t)=\{W^l(t)\}_{l=1,...,L}$ and current learning rate $\eta(t)$\\
        \hspace*{0.02in} {\bf Output:} 
        Updated weight $\mathcal{W}^{t+1}$ and updated learning rate $\eta^{t+1}$
        \begin{algorithmic}[1]
        \For{$l=1$ to $L$} 
            \State $\hat{Y}^{l}(t)$=OrderTwoBinaryConvolution$(X^l(t),W^l(t))$
        \EndFor
        \State $\frac{\partial{L}}{\partial{\widetilde{\mathcal{W}}}}=$BinaryBackward$(\frac{\partial{L}}{\partial{\hat{Y}}},\widetilde{\mathcal{W}})$
        \State $\mathcal{W}(t+1)=$UpdateParameters$(\mathcal{W}(t),\frac{\partial{L}}{\partial{\widetilde{\mathcal{W}}}},\eta(t))$
        \State
        $\eta(t+1)=$UpdateLearningrate$(\eta(t),t)$
        \end{algorithmic}
        \end{algorithm}

To train a HORQ-Net, we
quantize the input and weight filters and compute the binary convolution layer by layer. The binary convolution is detailed in Algorithm~\ref{SOBC}. After the Forward-pass, we use the binary weight $\widetilde{\mathcal{W}}$ and binary input $\widetilde{X}$ to do the back propagation. We also use the same way as Courbariaux \etal~\cite{courbariaux2016binarized} does to compute the gradient for the sign function $sign(\cdot)$. We should notice that we use the real-value weights and inputs when updating the parameters. The reason is that the parameter update is quite small in each iteration. If we update with binary weights, these updates may be eliminated during the binary operation in the next iteration and therefore the network will not be efficiently trained. The similar strategy is also applied in~\cite{courbariaux2015binaryconnect,courbariaux2016binarized,rastegari2016xnor}

\section{Experiments}
\label{section4}
In this section, we will show two main comparison experiments on MNIST and CIFAR-10. We compare HORQ-Net with some of the previous methods. Experiments show that HORQ-Net possesses better performance on image classification tasks.
\subsection{MNIST}
\label{setction4.1}

We test our HORQ-Net on MNIST dataset, which is a benchmark image classification dataset~\cite{lecun1998gradient} of handwritten digits from 0 to 9. To make this experiment comparable with BC~\cite{courbariaux2015binaryconnect} and BNN~\cite{courbariaux2016binarized}, we also use a MLP with a similar structure. This MLP consists of 3 hidden layers with 4096 Order-Two Residual Quantized connections and a L2-SVM layer with the Hinge loss (Lee \etal~\cite{DBLP:conf/aistats/LeeXGZT15} showed that L2-SVM is better than Softmax in this dataset). To train this MLP, we do not use any convolution, preprocessing, data-augmentation or pre-training skills. We use ADAM adaptive learning rate method~\cite{journals/corr/KingmaB14}. We use Batch Normalization with a minibatch of size 200 to speed up the training.

    \begin{figure}[t]
    \begin{center}
       \includegraphics[width=1\linewidth]{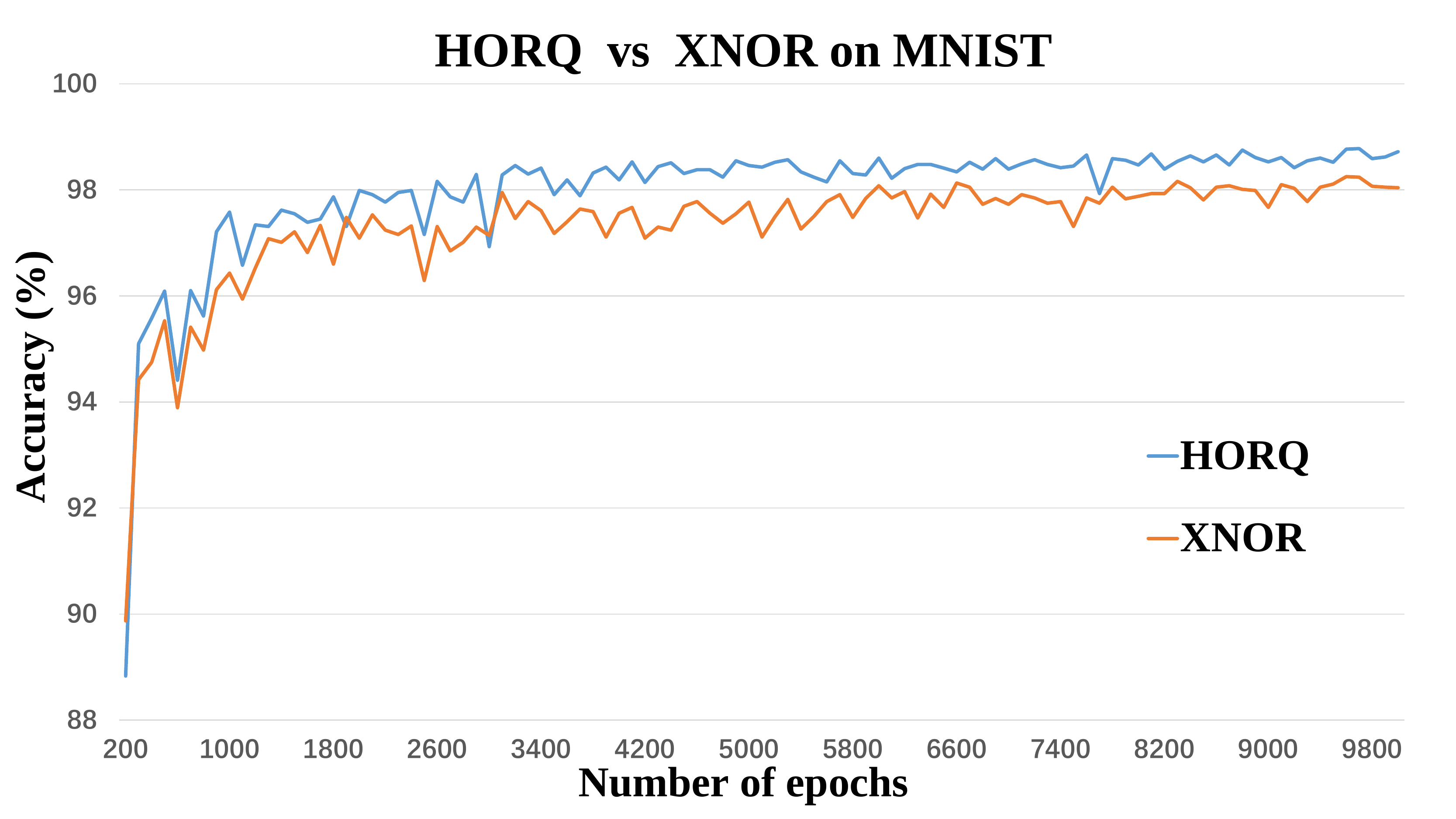}
    \end{center}
       \caption{This figure shows the classification accuracy of HORQ-Network and XNOR-Network on MNIST.}
       \label{MNISTaccuracy}
    \end{figure}
    \begin{figure}[t]
    \begin{center}
       \includegraphics[width=1\linewidth]{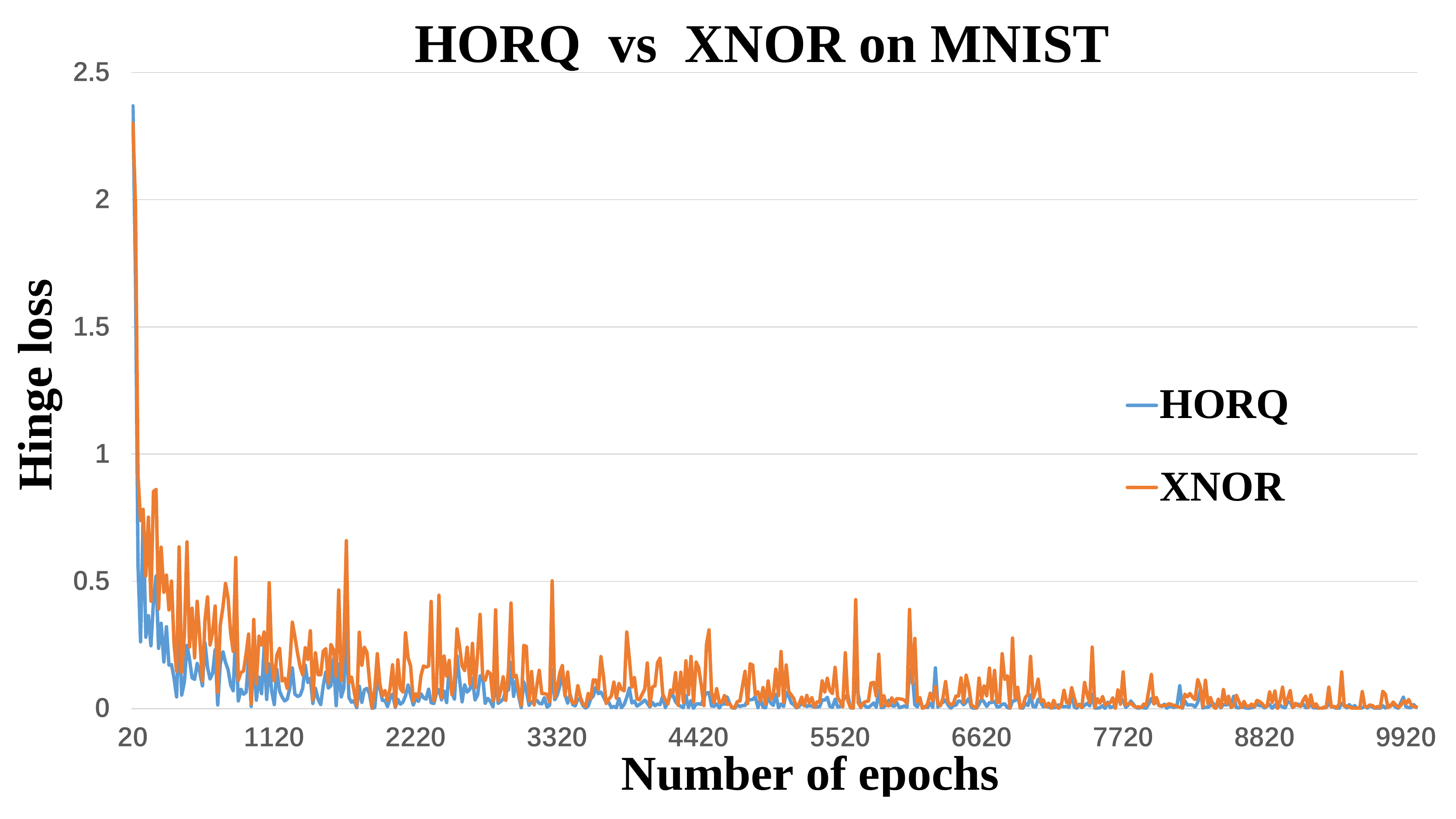}
    \end{center}
       \caption{This figure shows the hinge loss of HORQ-Network and XNOR-Network on MNIST.}
       \label{MNISTloss}
    \end{figure}

    \begin{table}[t]
    \begin{center}
    \begin{tabular}{cccc}
    \hlinew{1.5pt}
    Method & Binary Input& Binary Weight & Test error\\
    \hline
    BEB & No & Yes & 2.12\%  \\
    BC & No & Yes & 1.18\% \\
    BN & No & Yes & 0.96\%  \\
    BNN & Yes & Yes & 1.33\% \\
    XNOR & Yes & Yes & 1.96\%  \\
    \hline
    HORQ & Yes & Yes & 1.25\% \\
    \hline
    \end{tabular}
    \end{center}
    \caption{This Table shows the Test error rate of different binary method on MNIST:
    BEB (Binary expectation backpropagation~\cite{DBLP:journals/corr/ChengSML15}),
    BC (BinaryConnect~\cite{courbariaux2015binaryconnect}),
    BN (BinaryNet~\cite{courbariaux2016binarized}),
    BNN (Bitwise Neural Networks~\cite{DBLP:journals/corr/KimS16}),
    XNOR (XNOR-Networks~\cite{rastegari2016xnor}),
    HORQ (This work).}
    \label{MNIST}
    \end{table}

We also train a same MLP with only order-one binary connections (XNOR) to compare the final test accuracy. The results are shown in Figure~\ref{MNISTaccuracy} and Figure~\ref{MNISTloss}. We use the same network structure above to train XNOR-Net and HORQ-Net and find that HORQ-Net outperforms XNOR-Net by $0.71\%$ in accuracy. From Figure~\ref{MNISTaccuracy}, we also observe that HORQ-Net converges within fewer epochs. Figure~\ref{MNISTloss} shows the hinge loss changes over epoch. Both HORQ-Net and XNOR-Net can converge to a relatively small loss but the hinge loss curve of XNOR-Net is not as smooth as the loss curve of HORQ-Net. Most previous works (showed in Table~\ref{MNIST}) used binary weights and float-precision inputs. XNOR-Net~\cite{rastegari2016xnor} and our HORQ-Net use both binary weights and binary inputs. This experiment shows HORQ-Net can realize the acceleration of neural networks with little performance degradation.

\subsection{CIFAR-10}
\label{setction4.2}

We also test our HORQ-Network on CIFAR-10 dataset containing 50000 training images and 10000 testing images. We do not use any preprocessing or data-augmentation skills (which is showed to be a game changer in this data set~\cite{DBLP:journals/corr/Graham14}). In order to show the difference between the performance of methods using Order-Two Residual Quantization (HORQ) and order-one binary approximation (XNOR), firstly we use a shallow convolution neural network. The structure of our CNN is:
\begin{equation}
\begin{aligned}
  (32)C5-S-MP3-N-(32)C5-S-MP3-\\N-(64)C5-S-AP3-10FC-SOFTMAX
\end{aligned}
\end{equation}
Where $C5$ is a $5\times5$ convolution layer, S is a sigmoid activation layer, MP3 is a max-pooling layer with kernel size 3 and stride 2, AP3 is a average-pooling layer with kernel size 3 and stride 2, N is a LRN layers, FC is a fully connected layer and SOFTMAX is a softmax loss layer. To train this CNN, we set the size of the minibatch to 50 to speed up the training. We also centralize and standardize the training data.
\begin{figure}[t]
\begin{center}
   \includegraphics[width=1\linewidth]{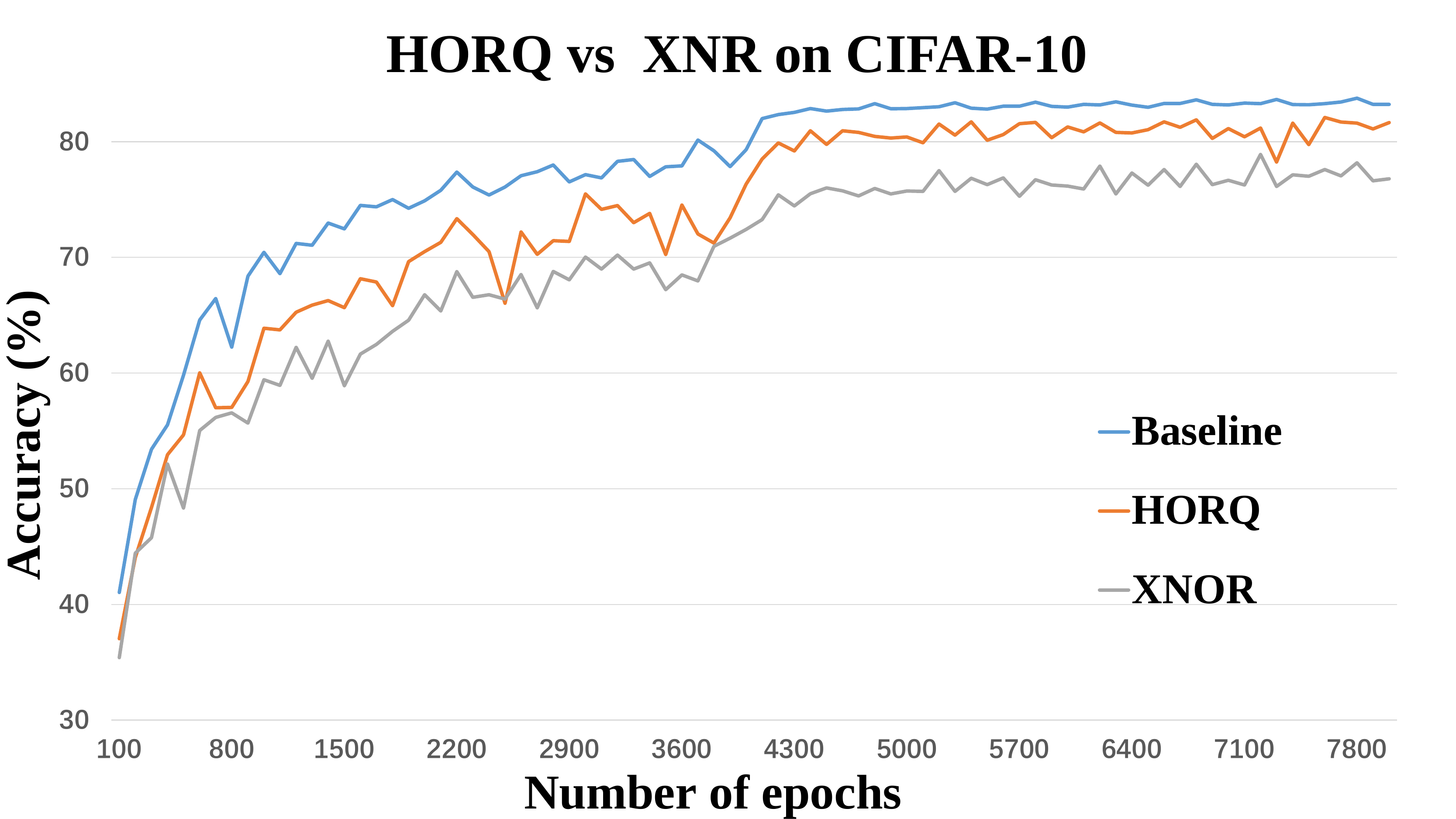}
\end{center}
   \caption{This figure shows the classification accuracy of HORQ-Network and XNOR-Network on CIFAR-10 on a shallow CNN.}
   \label{CIFARaccuracy}
\end{figure}
\begin{figure}[t]
\begin{center}
   \includegraphics[width=1\linewidth]{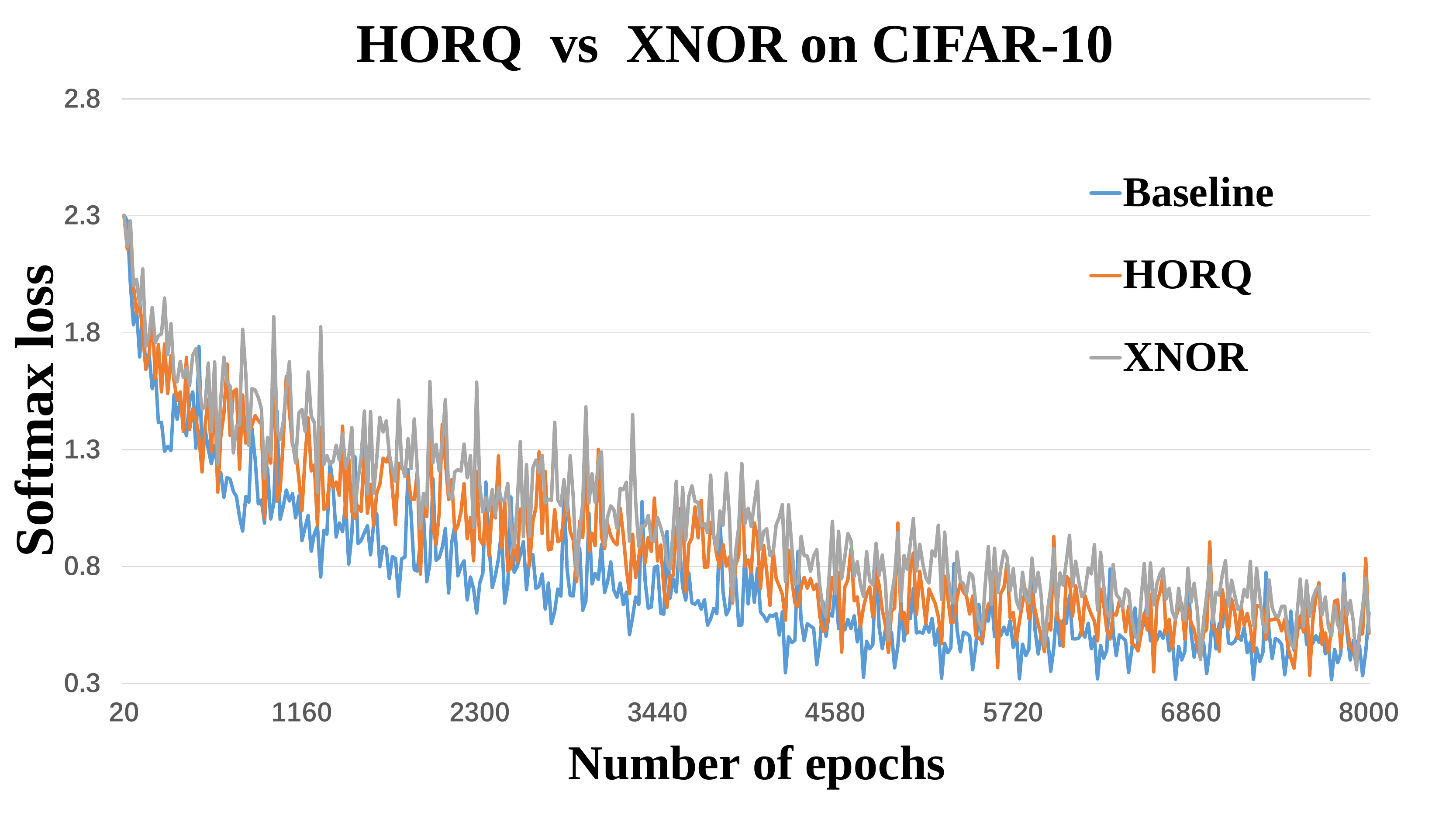}
\end{center}
   \caption{This figure shows the softmax loss of HORQ-Network and XNOR-Network on CIFAR-10 on a shallow CNN.}
   \label{CIFARloss}
\end{figure}

Since our CNN structure is not as complex as ConvNet~\cite{courbariaux2015binaryconnect} (ConvNet has six convolutional layers and two fully connected layers and each layer has more perceptions), our baseline (without using any binary approximation) accuracy is not as high as theirs. But this shallow network makes it easier to compare the performance between HORQ and XNOR under the same initialization, parameter setting and training strategy. We report the final performance in Figure~\ref{CIFARaccuracy} and Figure~\ref{CIFARloss}. Using the same network structure, HORQ-Net converges with accuracy drop within $2\%$ compared with our baseline. The accuracy drops $\sim5\%$ in XNOR-Net. Besides, HORQ-Net and XNOR-Net converges in a similar speed. Hence this experiment also shows the better performance of HORQ-Net.

\subsection{Storage Space Analysis}
\label{section4.3}

\begin{figure}[h]
\begin{center}
   \includegraphics[width=1\linewidth]{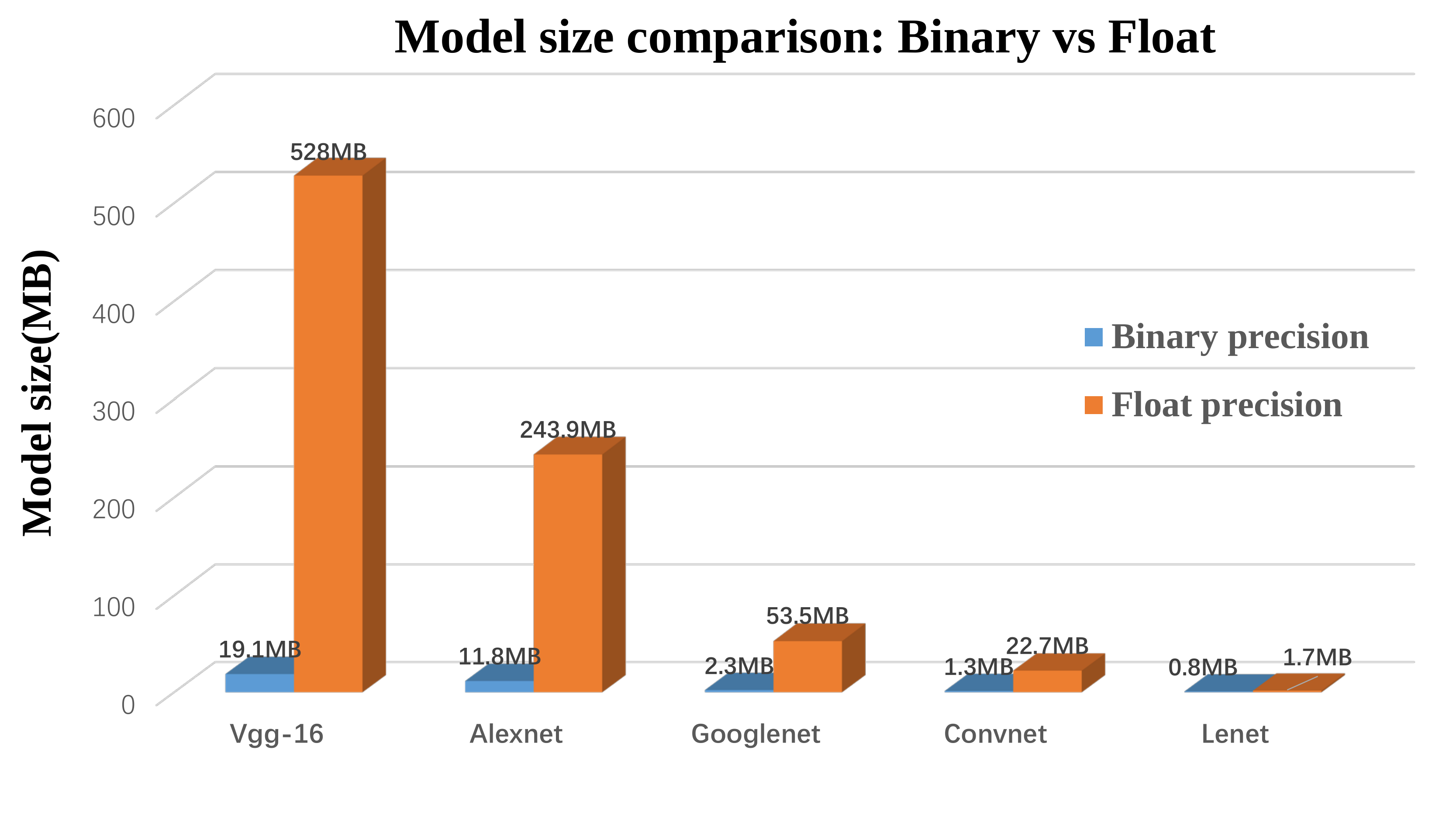}
\end{center}
   \caption{This figure lists some models(Vgg-16~\cite{Simonyan2014Very}, Alexnet~\cite{krizhevsky2012imagenet}, Googlenet~\cite{DBLP:conf/cvpr/SzegedyLJSRAEVR15}, Convnet~\cite{courbariaux2015binaryconnect}, Lenet~\cite{Haykin2009GradientBased})
   shows the Comparison of storage space of several models between float precision and binary precision.}
   \label{storage}
\end{figure}

Generally speaking, our high-order binarization can be applied to any DCNN models. Models with binary weights will take up less storage memory than models with double precision weights. A very deep convolutional neural networks, for example, VGG-16, will occupy nearly 400M storage space using float precision. Figure~\ref{storage} shows the storage cost of some widely used models with double and binary precision weights.

\subsection{Computation Analysis}
\label{section4.4}

        \begin{figure}[t]
        \begin{center}
           \includegraphics[width=1\linewidth]{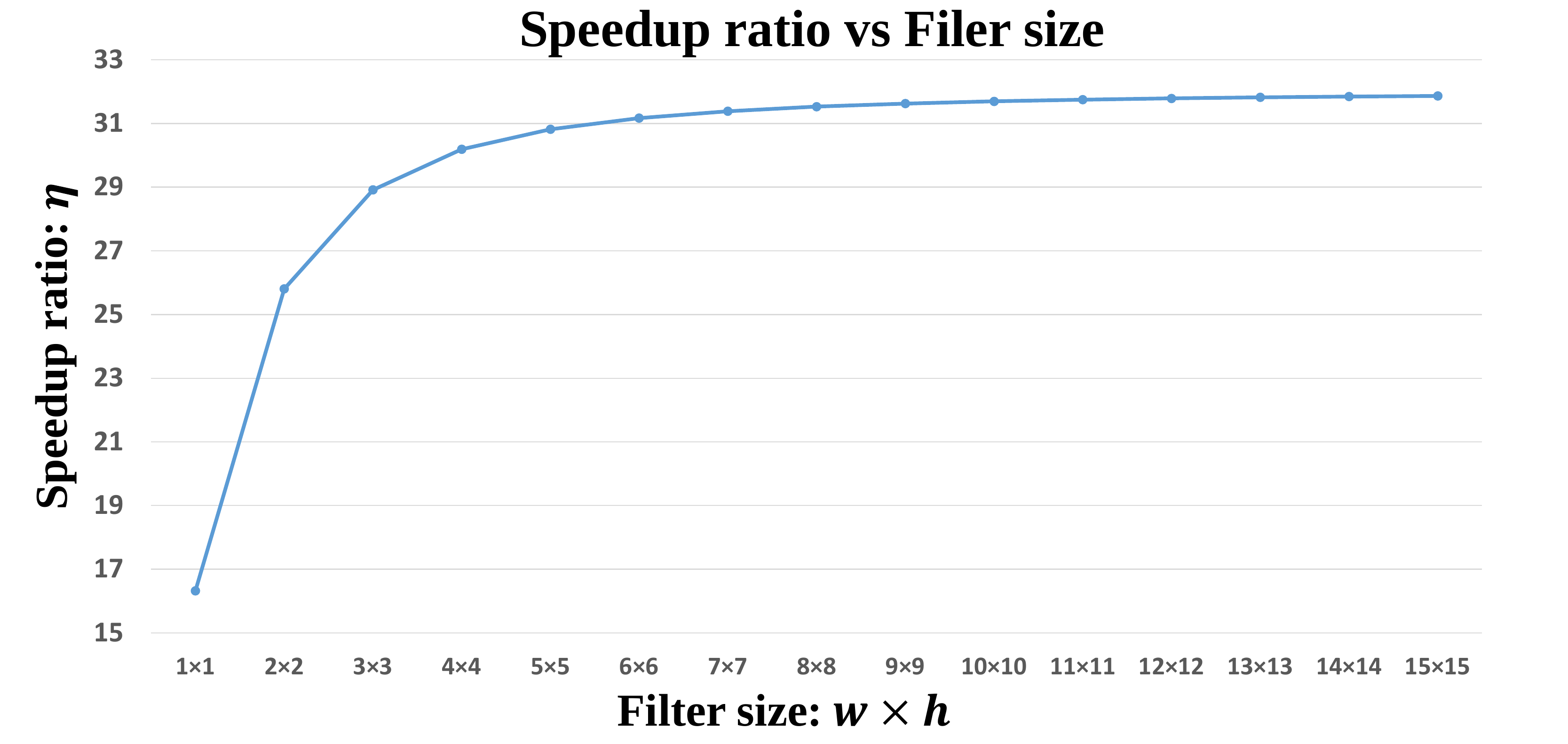}
           \includegraphics[width=1\linewidth]{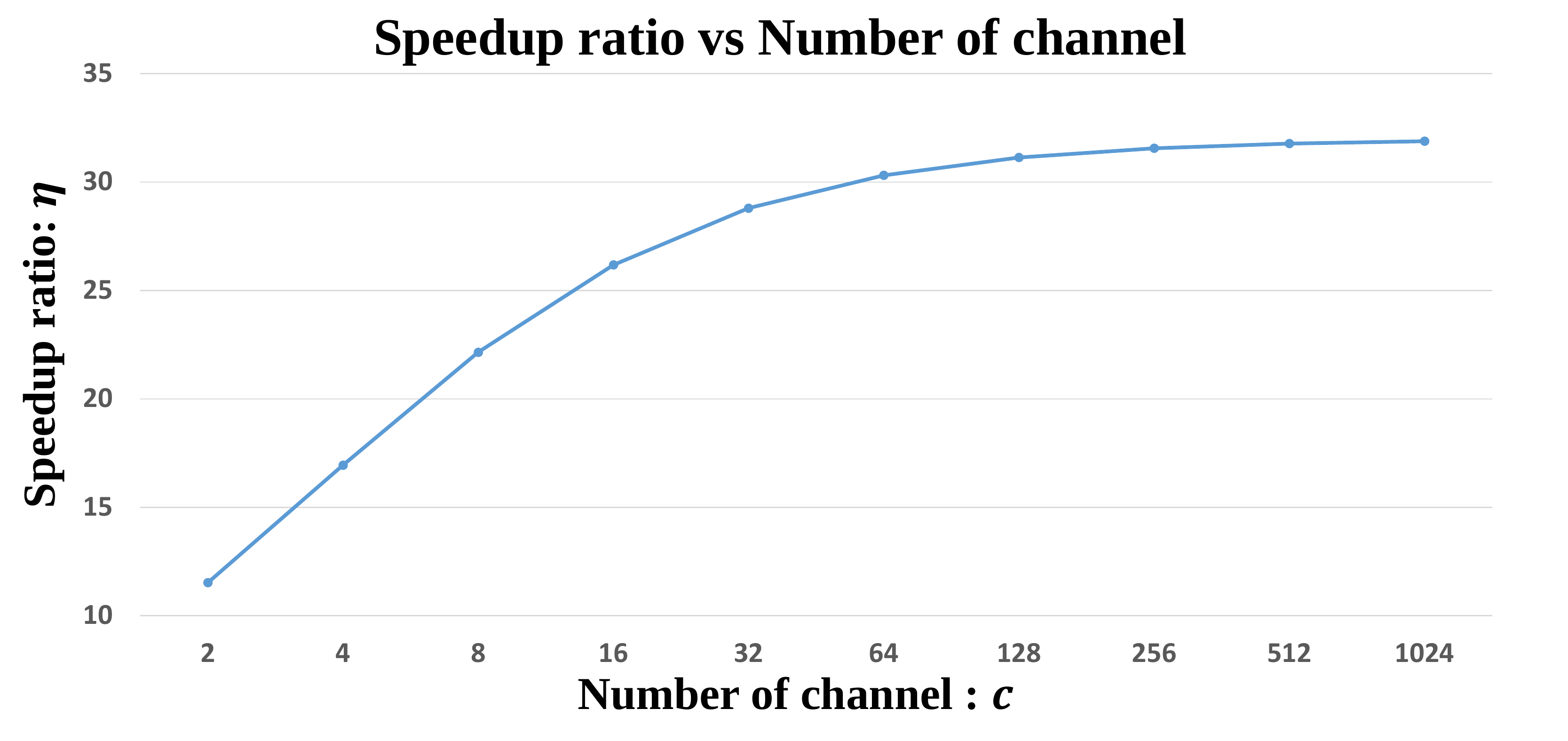}
           \includegraphics[width=1\linewidth]{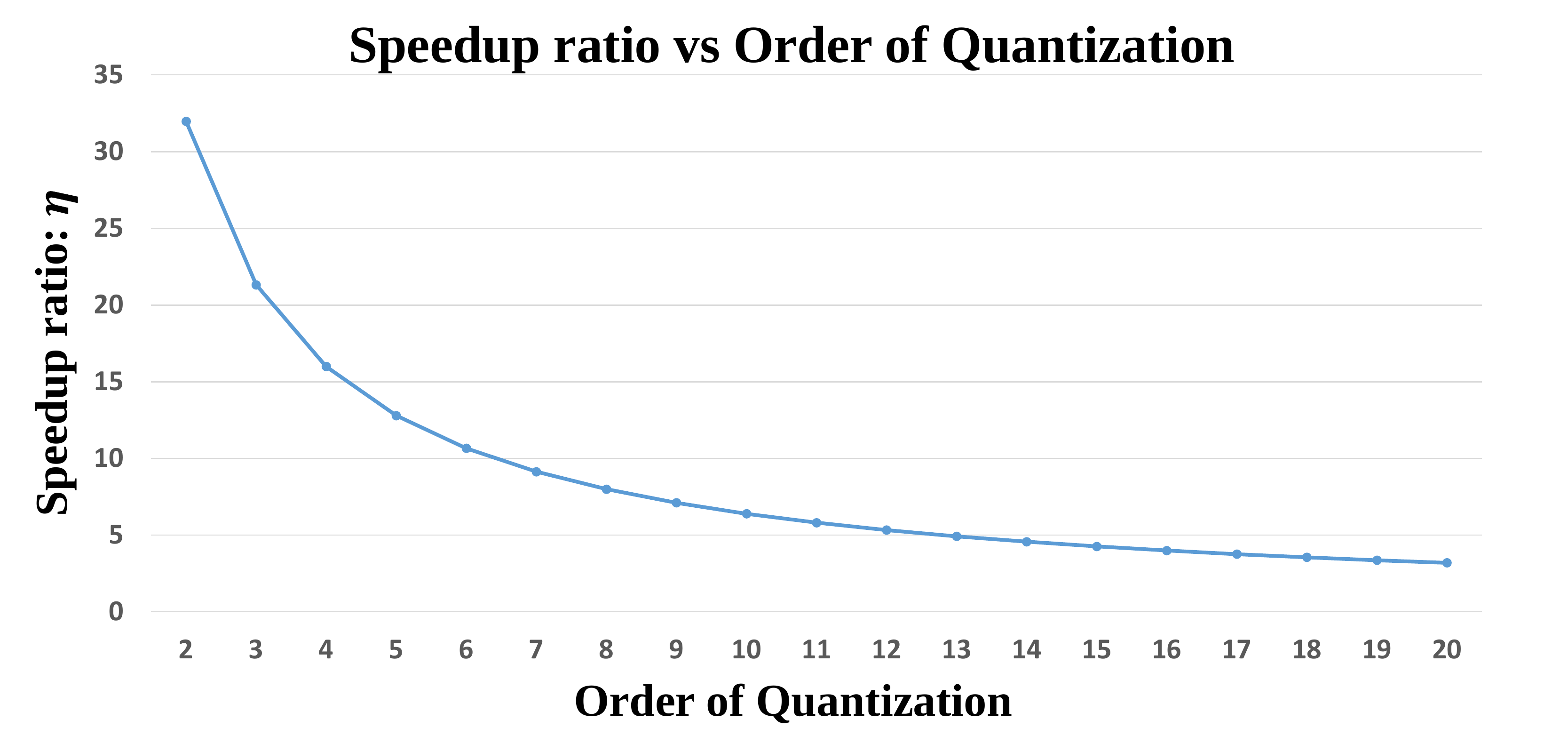}
        \end{center}
           \caption{This figure shows the relationship between (a)speedup ratio and filter size, (b)speedup ratio and channels, (c)speedup ratio and order of quantization.}
        \label{eta_vs_wh}
        \label{eta_vs_c}
        \label{eta_vs_o}
        \end{figure}


Consider a convolution operation $(I,W,*)$, where input $I\in\mathbb{R}^{c_{in}\times w_{in}\times h_{in}}$, Weight tensor $W\in\mathbb{R}^{c_{out}\times c_{in}\times w\times h}$, the total number of operation is $c_{out}\times c_{in}\times wh\times w_{in}h_{in}$. Using the current generation CPU, which is capable of performing 64 binary operations within one cycle clock, our method of High-Order Residual Quantization of order K needs $K\times c_{out}\times c_{in}\times wh\times w_{in}h_{in}+(K+1)\times w_{in}h_{in}=KN_p+(K+1)N_n$ operations. Among these operations, $KN_p$ operations are binary-precision operations, which can be sped up, while the other $(K+1)N_n$ operations are float-precision operations, which cannot be sped up. Thus the speedup ratio can be computed as:
\begin{equation}
\label{speedupratio}
\begin{aligned}
  \eta
  &=\frac{c_{out}c_{in}whw_{in}h_{in}}{\frac{1}{64}(Kc_{out}c_{in}whw_{in}h_{in})+(K+1)w_{in}h_{in}}\\
  &=\frac{64c_{out}c_{in}wh}{Kc_{out}c_{in}wh+64(K+1)}
\end{aligned}
\end{equation}
For the case of Order-Two, we can compute the speedup ratio:
\begin{equation}
\label{speedupratio}
\begin{aligned}
  \eta
  &=\frac{c_{out}c_{in}whw_{in}h_{in}}{\frac{1}{64}(2c_{out}c_{in}whw_{in}h_{in})+3w_{in}h_{in}}\\
  &=\frac{64c_{out}c_{in}wh}{2c_{out}c_{in}wh+192}
\end{aligned}
\end{equation}

\begin{table}[t]
\begin{center}
\begin{tabular}{ll}
\hlinew{1.5pt}
Method & Speedup ratio\\
\hline
Order-One Residual Quantization(XNOR) & $58\times$  \\
Order-Two Residual Quantization & $30\times$  \\
Order-Three Residual Quantization & $20\times$  \\
Order-Four Residual Quantization & $15\times$  \\
\hline
\end{tabular}
\end{center}
\caption{This table shows speedup ratio using HORQ method in different orders. XNOR-Net can be considered as Order-One Residual Quantization.}
\label{MNIST}
\end{table}

As we can see in Equation~\ref{speedupratio}, the speedup ratio does not depend on the width or the height of the input tensor but on the filter size: $wh$ and the number of channels: $c_{in}c_{out}$. Firstly, we fix the number of channels: $c_{in}c_{out}=10\times 10$ to see how filter size influence speedup ratio. Secondly, we fix the filter size : $w\times h=3\times 3$ and input channels $c_{in}=3$ to see how output channels influence speedup ratio. As we can see from Figure~\ref{eta_vs_c}, the speedup will not be remarkable if the number of channels and filter size is two small. Thus when we apply the binary method to DCNN, we should avoid quantizing layers with few channels (e.g. first layer with 3 channels). If we set $c_{in}c_{out}=64\times 256, w\times h=3\times 3$, our Order-Two Residual Quantization can reach $31.98\times$ speedup. But in practice, the speedup ratio may be a little bit lower due to the process of memory read and data preprocessing. From Figure~\ref{eta_vs_o}, we observe that Order-Two and Order-Three Residual Quantization still remain a relatively high speed up ratio ($>20\times$). Thus our HORQ method of order-two and order-three are very powerful in accelerating the neural network with performance guaranteed.

\section{Conclusion}

In this paper, we propose an efficient and accurate binary approximation method called High-Order Residual Quantization. We introduce the concept of residual to represent the information loss and recursively compute the quantized residual to reduce the information loss. Using binary weights, the size of network is reduced by $\sim 32\times$ and this method provides $\sim 30\times$ speed up. This also provides the possibility of running the inference of deep convolutional network on CPU. Our experiments show that the performance of HORQ-net is guaranteed. HORQ-Net outperforms XNOR-Net in MNIST($0.71\%$) and in CIFAR-10($\sim3\%$).

\section{Acknowledgements}
The work was supported by State Key Research and Development Program (2016YFB1001003). This work was also supported by NSFC (U1611461, 61502301), China's Thousand Youth Talents Plan, the 111 Program, the Shanghai Key Laboratory of Digital Media Processing and Transmissions, and Cooperative Medianet Innovation Center.

\newpage
\newpage
{\small
\bibliographystyle{ieee}
\bibliography{egbib1}
}

\end{document}